\documentclass[10pt,twocolumn,letterpaper]{article}

\usepackage{cvpr}
\usepackage{times}
\usepackage{epsfig}
\usepackage{eso-pic}

\usepackage[utf8]{inputenc} 
\usepackage[T1]{fontenc}    
\usepackage{url}            
\usepackage{booktabs}       
\usepackage{amsfonts}       
\usepackage{nicefrac}       
\usepackage{microtype}      

\usepackage{epstopdf}
\usepackage{algorithm}
\usepackage{algorithmic}
\usepackage{bm, amsfonts, amssymb, multirow, nicefrac, xr}
\usepackage{amsmath}
\usepackage{theorem}
\usepackage{caption}
\usepackage{graphicx}
\usepackage{caption}
\usepackage{subcaption}
\usepackage{enumerate}

\def\log{\mathrm{log}}
\newtheorem{theorem}{Theorem}

\newcommand{\defEq}{\stackrel{.}{=}}
\newcommand{\Real}{\mathbb{R}}
\newcommand{\yCorrupt}{\tilde{\bm{y}}}
\newcommand{\ellRobust}{\bm{\ell}^{\leftarrow}}
\newcommand{\ellRobustScalar}{\ell^{\leftarrow}}
\newcommand{\ellFb}{\bm{\ell}^{\rightarrow}}
\newcommand{\ellFbScalar}{\ell^{\rightarrow}}
\newcommand{\argmin}{\operatornamewithlimits{argmin}}
\newcommand{\argmax}{\operatornamewithlimits{argmax}}

\makeatletter
\newcommand{\specialcell}[1]{\ifmeasuring@#1\else\omit$\displaystyle#1$\ignorespaces\fi}
\makeatother

\usepackage[breaklinks=true,bookmarks=false]{hyperref}

\cvprfinalcopy 

\def\cvprPaperID{720} 

\begin{document}

\title{Making Deep Neural Networks Robust to Label Noise: \\ a Loss Correction Approach}

\author{First Author\\
Institution1\\
Institution1 address\\
{\tt\small firstauthor@i1.org}
\and
Second Author\\
Institution2\\
First line of institution2 address\\
{\tt\small secondauthor@i2.org}
}

\author{
  Giorgio Patrini$^{1,2}$, Alessandro Rozza$^{3}$, Aditya Krishna Menon$^{2, 1}$, Richard Nock$^{2, 1, 4}$, Lizhen Qu$^{2, 1}$\\\
  $^{1}$Australian National University, $^{2}$Data61, $^{3}$Waynaut, $^{4}$University of Sydney \\
{\tt\small \{name.surname\}@data61.csiro.au, alessandro.rozza@waynaut.com}
  }

\maketitle

\begin{abstract}
We present a theoretically grounded approach to train deep neural networks, including recurrent networks, subject to class-dependent label noise.
We propose two procedures for loss correction that are agnostic to both application domain and network architecture. They simply amount to at most a matrix inversion and multiplication, provided that we know the probability of each class being corrupted into another.
We further show how one can \emph{estimate} these probabilities, adapting a recent technique for noise estimation to the multi-class setting, and thus providing an end-to-end framework.
Extensive experiments on MNIST, IMDB, CIFAR-10, CIFAR-100
and a large scale dataset of clothing images
employing a diversity of architectures --- stacking dense, convolutional, pooling, dropout, batch normalization, word embedding, LSTM and residual layers --- demonstrate the noise robustness of our proposals. Incidentally, we also prove that, when ReLU is the only non-linearity, the loss curvature is immune to class-dependent label noise.

\end{abstract}

\section{Introduction}

Large datasets used in training modern machine learning models, such as deep neural networks, are often affected by label noise.
The problem is pervasive for a simple reason:
manual expert-labelling of each instance at a large scale is not feasible,
and so researchers often resort to cheap but imperfect surrogates.
Two such popular surrogates are crowdsourcing using non-expert labellers
and --- especially for images ---
the use of search engines to query instances by a keyword, assuming the keyword as a valid label \cite{ffpzLO, sczHI, dfgLE, nlxVR, kshtdpfTU}
Both approaches offer the possibility to scale the acquisition of training labels,
but
invariably result in the introduction of label noise,
which may adversely affect model training.

Our goal is to effectively train deep neural networks with modern architectures under label noise.
We do so by marrying two different lines of recent research.
The first strand is work on \emph{ad-hoc deep architectures} tailored to the problem, primarily developed
in Computer Vision~\cite{mhLT, rlazerTD, sbpbfTC, xxtwLF}.
While some such approaches have shown good experimental performance on specific domains, they lack a solid theoretical framework
and often need a large amount of clean labels to obtain acceptable results --- in particular, for pre-training or validating hyper-parameters~\cite{xxtwLF, kshtdpfTU, rlazerTD}.

The second strand is recent Machine Learning research on \emph{theoretically grounded means of combating label noise}.
In particular, we are interested in the design of \emph{corrected losses} that are \emph{robust} to label noise \cite{srLS, ndrtLW, pnncLF}. Despite their formal guarantees, these methods have not been fully appreciated in practice because, crucially, they require noise rates to be known \emph{a priori}.

An estimate of the noise is often available to practitioners by polishing a subset of the training data \cite{xxtwLF} --- which is useful and often necessary for model selection.
Yet, interestingly, recent work has provided practical algorithms for estimating the noise rates \cite{Scott:2013, Sanderson:2014, ltCW, mvrowLF, Ramaswamy:2016};
remarkably, this is achievable with \emph{absolutely no knowledge of ground truth labels}.
To our knowledge, no prior work has combined those estimators with loss correction techniques,
nor has either idea been applied to modern deep architectures.
Our contributions aim to unify these research streams:
\begin{itemize}
\item We introduce two alternative procedures for loss correction, provided that we know a stochastic matrix $T$ summarizing the probability of one class being flipped into another under noise. The first procedure, a multi-class extension of \cite{ndrtLW, pnncLF} applied to neural networks, is called ``\textit{backward}'' as it multiplies the loss by $T^{-1}$. The second, inspired by \cite{sbpbfTC}, is named ``\textit{forward}'' as it multiplies the network predictions by $T$.
\item We prove that both procedures enjoy formal robustness guarantees \emph{w.r.t.} the clean data distribution. Since we only operate on the loss function, the approach is both architecture and application domain independent, as well as viable for any chosen loss function.
\item We take a further step and extend the noise estimator of \cite{mvrowLF} to our multi-class setting, thus formulating an end-to-end solution to the problem. 
\item We prove that for ReLU networks the Hessian of the loss is independent from label noise.
\end{itemize}

We apply our loss corrections to image recognition on MNIST, CIFAR-10, CIFAR-100 and sentiment analysis on IMDB; we simulate corruption by artificially injecting noise on the training labels. In order to show that no architectural choice is the secret ingredient of our robustification recipe, we experiment with a variety of network modules currently in fashion: convolutions and pooling \cite{lbbhGB}, dropout \cite{shkssDA}, batch normalization \cite{isBN}, word embedding and residual units \cite{He_2015, hzrsIM}. 
Additional tests on LSTM \cite{hsLS} confirm that the procedures can be seamlessly applied to recurrent neural networks as well. Comparisons with non-corrected losses and several known methods confirm robustness of our two procedures, with the forward correction dominating the backward. Unsurprisingly, the noise estimator is the bottleneck in obtaining near-perfect robustness, yet in most experiments our approach is often the best compared to prior work. Finally, we experiment with Clothing1M, the $1M$ clothing images dataset of \cite{xxtwLF}, and establish the new state of the art.


\section{Related work}\label{sec:rel}

Our work leverages recent research in a number of different areas, summarized below.

\emph{Noise robustness\footnote{We use the term robustness in its meaning of immunity to noise and not generically as ``adaptivity to various scenarios'', \emph{e.g.} \cite{fkesbhEA}.}.}
Learning with noisy labels has been widely investigated in the literature \cite{Frenay_2014}.
From the theoretical standpoint label noise has been studied in two different regimes, with vastly different conclusions.
In the case of low-capacity (typically linear) models, even mild symmetric, \emph{i.e.} class-independent (versus asymmetric, \emph{i.e.} class-dependent), label noise can produce solutions that are akin to random guessing \cite{lsRC}.
On the other hand, the Bayes-optimal classifier remains unchanged under symmetric \cite{ndrtLW,mvrowLF} and even instance dependent label noise \cite{mrnBL} implying that high-capacity models are robust to essentially any level of such noise, given sufficiently many samples. 

\emph{Surrogate losses}.
Suppose one wishes to minimize a loss $\ell$ on clean data.
When the level of noise is known \emph{a priori},
\cite{ndrtLW} provided the general form of a \emph{noise corrected} loss $\hat{\ell}$
such that minimization of $\hat{\ell}$ on noisy data
is \emph{equivalent}
to minimization of ${\ell}$ on clean data.
In the idealized case of symmetric label noise, for certain $\ell$ one in fact does not need to know the noise rate:
\cite{gmsMR} gives a sufficient condition for which $\ell$ is robust, and several examples of such robust non-convex losses,
while \cite{rmwLW} shows that the (convex) linear or \emph{unhinged} loss is its own noise-corrected loss. Another robust non-convex loss is given in \cite{msvOT}.

\emph{Noise rate estimation}.
Recent work has provided methods to estimate label flip probabilities directly from noisy samples.
Typically, it is required that the generating distribution is 
such that for each class, there exists some ``perfect'' instance, \emph{i.e.} one that is classified with probability equal to one.
Proposed estimators involve either the use of kernel mean embedding \cite{Ramaswamy:2016},
or post-processing the output of a standard class-probability estimator such as logistic regression 
using order statistics on the range of scores \cite{ltCW, mvrowLF}
or the slope of the induced ROC curve \cite{Sanderson:2014}.

\emph{Deep learning with noisy labels}.
Several works in Deep Learning have attempted to deal with noisy labels of late, especially in Computer Vision. This is often achieved by formulating noise-aware models. \cite{mhLT} builds a noise model for binary classification of aerial image patches, which can handle omission and wrong location of training labels. 
\cite{xxtwLF} constructs a more sophisticated mix of symmetric, asymmetric and instance-dependent noise; two networks are learned by EM as models for classifier and noise type. 
It is often the case that a small set of clean labels is needed in order either to pre-train or fine-tune the model \cite{xxtwLF, kshtdpfTU, rlazerTD}.

The work of \cite{sbpbfTC} deserves a particular mention. The method augments the architecture by adding a linear layer on top of the network. Once learned, this layer plays the role of our matrix $T$. 
However, learning 
this architecture appears problematic; heuristics such as trace regularization and a fixed updating schedule for the linear layer are necessary. We sidestep those issues by decoupling the two phases: we first estimate $T$ and then learn with loss correction.



We are not aware of any other attempt at either applying the noise-corrected loss approach of \cite{ndrtLW} to neural networks,
nor on combining those losses with the above noise rate estimators.
Our work sits precisely in this intersection.
Note that, even though in principle loss correction should not be necessary for high-capacity models like deep neural networks, owing to aforementioned theoretical results,
in practice, such correction may offset the sub-optimality of these models arising from training on finite samples.
Specifically,
we expect that directly optimizing the (corrected) objective we care about will be beneficial in the finite-sample case. 

\section{Preliminaries}\label{sec:setting}

We begin by fixing notation.
We let $[c] \defEq \{1,\dots, c\}$ for any $c$ positive integer.
Column vectors are written in bold (\eg $\bm{v}$) and matrices in capitals (\eg $V$).
Coordinates of a vector are denoted by a subscript (\eg $\bm{v}_j$), while rows and columns of a matrix are denoted \eg $V_{j \cdot}$ and $V_{\cdot j}$ respectively. We denote the all-ones vector by $\bm{1}$, with size clear from context, and $\Delta^{c-1} \subset [0,1]^c$ the $c$-dimensional simplex.

In supervised $c$-class classification, 
one has feature space $\mathcal{X} \subseteq \mathbb{R}^d$
and label space $\mathcal{Y} = \{ \bm{e}^i: i \in [ c ] \}$,
where $\bm{e}^i$
denotes the $i$th standard canonical vector in $\Real^c$ by , \ie $\bm{e}^i \in \{0,1 \}^c, \bm{1}^\top \bm{e}^i = 1$.
One observes examples $(\bm{x}, \bm{y})$ drawn from 
an unknown distribution $p(\bm{x}, \bm{y}) = p(\bm{y}|\bm{x}) p(\bm{x})$ over $\mathcal{X} \times \mathcal{Y}$. We denote expectations over $p(\bm{x}, \bm{y})$ by $\mathbb{E}_{\bm{x}, \bm{y}}$.
Note that each $\bm{y}$ only has one non-zero value at the coordinate corresponding to the underlying label.

An $n$-layer neural network\footnote{\emph{W.l.o.g.}, we assume all layers to be \emph{fully connected}, or dense; for example, convolutions can be represented by dense layers with shared sparse weights.} comprises a transformation $\bm{h} \colon \mathcal{X} \to \Real^c$,
where $\bm{h} = (\bm{h}^{(n)} \circ \bm{h}^{(n-1)} \circ \dots \circ \bm{h}^{(1)})$
is the composition of a number of intermediate transformations --- the layers --- defined by:
\begin{alignat*}{2}
	( \forall i \in [ n - 1 ] ) \, &\bm{h}^{(i)}( \bm{z} ) &&= \bm{\sigma}(W^{(i)}\bm{z} + \bm{b}^{(i)})\:\:, \\
	&\bm{h}^{(n)}( \bm{z} ) &&= W^{(i)}\bm{z} + \bm{b}^{(i)}\:\:.
\end{alignat*}
where
$W^{(i)} \in \mathbb{R}^{d^{(i)} \times d^{(i - 1)}}$ and $\bm{b}^{(i)} \in \mathbb{R}^{d^{(i)}}$ are parameters to be estimated\footnote{Here, $d^{(0)} = d$, the original feature dimensionality, and $d^{(n)} = c$, the label dimensionality.},
and
$\bm{\sigma}$ is any activation function that acts \emph{coordinate-wise},
such as the ReLU $\bm{\sigma}(\bm{x})_i = \max(0, \bm{x}_i)$.
Observe that the final layer applies a \emph{linear} projection, unlike all preceding layers.
To simplify notation, we write:
$$ ( \forall i \in [ n ] ) \, \bm{x}^{(i)} \defEq \bm{h}^{(i)}( \bm{x}^{(i - 1)} ), $$
with the base case $\bm{x}^{(0)} \defEq \bm{x}$,
so that \eg $\bm{x}^{(1)}$ is exactly the representation in the first layer. The coordinates of $\bm{h}( \bm{x} )$ represent the relative weights that the model assigns to each class $i = 1, \dots, c$ to be predicted. The predicted label is thus given by $\arg \max_{i \in [c]} \bm{h}_i(\bm{x})$. In the training phase, the output of the final layer is contrasted with the true label $\bm{y}$ via two steps.
First, $\bm{h}(\cdot)$ passes through the \emph{softmax} function
$e^{\bm{h}_i(\bm{x})} / \sum_{k=1}^{c} e^{\bm{h}_k(\bm{x})} $.
The softmax output can be interpreted as a vector approximating the class-conditional probabilities $p(\bm{y}|\bm{x})$; we denote it by $\hat{p}(\bm{y}|\bm{x}) \in \Delta^{c-1}$.
Next, we measure the discrepancy between label $\bm{y} = \bm{e}^i$ and network output by a loss function $\ell \colon \mathcal{Y} \times \Delta^{c-1} \to \Real$, for example by means of \emph{cross-entropy}:
\begin{align}
\ell(\bm{e}^i, \hat{p}(\bm{y}|\bm{x})) = - (\bm{e}^i )^\top \log ~ \hat{p}(\bm{y}|\bm{x}) = - \log ~ \hat{p}(\bm{y} = \bm{e}^i|\bm{x}) \:\:.
\end{align}
\noindent With some abuse of notation, we also define a loss in vector form $\bm{\ell}: \Delta^{c-1} \rightarrow \Real^c$, computed on every possible label:
\begin{align}
\bm{\ell} (\hat{p}(\bm{y}|\bm{x})) = \left(\ell(\bm{e}^1, \hat{p}(\bm{y}|\bm{x})), \dots, \ell(\bm{e}^c, \hat{p}(\bm{y}|\bm{x})) \right)^\top \in \Real^{c} \:\:.
\end{align}
In the following, formal results hold under very mild conditions on a generic loss function $\ell$; at times we provide examples for the cross-entropy. For simplicity, one could think of cross-entropy every time $\ell$ is mentioned.

\section{Label noise and loss robustness}
\label{sec:noise}

We now consider label noise.
We assume the \emph{asymmetric}, \emph{i.e.} class-conditional noise setting \cite{ndrtLW},
where
each label $\bm{y}$ in the training set
is flipped to $\yCorrupt \in \mathcal{Y}$
with probability $p(\yCorrupt | \bm{y})$;
feature vectors are untouched.
Thus, we observe samples from a distribution $p(\bm{x}, \yCorrupt) = \sum_{\bm{y}} p(\yCorrupt | \bm{y}) p(\bm{y}|\bm{x}) p(\bm{x})$.
Denote by $T \in [ 0, 1 ]^{c \times c}$ the noise transition matrix specifying the probability of one label being flipped to another,
so that $\forall i,j\:\:~T_{i j} = p(\yCorrupt=\bm{e}^j | \bm{y}=\bm{e}^i)$.
The matrix is row-stochastic and not necessarily symmetric across the classes.

This is an approximation of real-world corruption which can still be useful in certain scenarios. One such case is that of classes representing a fine-grained hierarchy of concepts, for example dog breeds and bird species \cite{kshtdpfTU} or narrow categories of clothing \cite{xxtwLF}. Classes may be too similar between each other for non-expert human labellers to distinguish, regardless of the specific instances. Little is known about learning under the more generic \emph{feature dependent} noise, with few exceptions \cite{xxtwLF, gmsMR, mrnBL}.

We aim to modify a loss $\ell$ so as to make it robust to asymmetric label noise; in fact, this is possible if $T$ is known. Under this assumption --- that we relax later on --- we introduce two alternative corrections inspired by \cite{ndrtLW} and \cite{sbpbfTC}.

\subsection{The backward correction procedure}

We can build an \emph{unbiased estimator} of the loss function, such that \emph{under expected label noise} the corrected loss equals the original one computed on clean data. This property is stated in the next Theorem, a multi-class generalization of \cite[Theorem 1]{ndrtLW}. The Theorem is also a particular instance of the more abstract \cite[Theorem 3.2]{vML}.

\begin{theorem}\label{th:unbias}
Suppose that the noise matrix $T$ is non-singular.
Given a loss $\bm{\ell}$, \emph{backward} corrected loss is defined as:
$$
\ellRobust(\hat{p}(\bm{y}|\bm{x})) = T^{-1} \bm{\ell}(\hat{p}(\bm{y}|\bm{x}))\:\:.$$
Then, the loss correction is unbiased, \emph{i.e.} :
$$
\forall \bm{x},~~\, \mathbb{E}_{\tilde{\bm{y}} | \bm{x}} ~\ellRobust(\bm{y}, \hat{p}(\bm{y}|\bm{x})) 
= \mathbb{E}_{\bm{y} | \bm{x}}~\bm{\ell}(\bm{y}, \hat{p}(\bm{y}|\bm{x}))\:\:,
$$
\noindent and therefore the minimizers are the same:
$$
\argmin_{\hat{p}(\bm{y}|\bm{x})} \mathbb{E}_{\bm{x}, \tilde{\bm{y}}}~\ellRobustScalar(\bm{y}, \hat{p}(\bm{y}|\bm{x})) = 
\argmin_{\hat{p}(\bm{y}|\bm{x})} \mathbb{E}_{\bm{x}, \bm{y}}~\ell(\bm{y}, \hat{p}(\bm{y}|\bm{x}))\:\:.
$$
\end{theorem}
\noindent \textbf{Proof}.
$
\mathbb{E}_{\yCorrupt | \bm{x}}~\ellRobust(\hat{p}(\bm{y}|\bm{x}))
= \mathbb{E}_{\bm{y} | \bm{x}}~T \ellRobust(\hat{p}(\bm{y}|\bm{x}))
= \mathbb{E}_{\bm{y} | \bm{x}}~T~T^{-1} \bm{\ell}(\hat{p}(\bm{y}|\bm{x}))
= \mathbb{E}_{\bm{y} | \bm{x}}~\bm{\ell}(\hat{p}(\bm{y}|\bm{x}))\:\:.$ \hfill $\blacksquare$ \\

The corrected loss is effectively a linear combination of the loss values for each observable label, whose coefficients are due to the probability that $T^{-1}$ attributes to each possible true label $\bm{y}$, given the observed one $\yCorrupt$. Intuitively, we are ``going one step back'' in the noise process described by the Markov chain $T$.
The corrected loss is differentiable --- although not always non-negative --- and can be minimized with any off-the-shelf algorithm for back-propagation. Although in practice $T$ would be invertible almost surely, its condition number may be problematic. A simple solution is to mix $T$ with the identity matrix before inversion; this may be seen as taking a more conservative noise-free prior.

\subsection{The forward correction procedure}
Alternatively, we can correct the model predictions. Following \cite{sbpbfTC}, we start by observing that a neural network learned with no loss correction would result in a predictor for noisy labels $\hat{p}(\tilde{\bm{y}}|\bm{x})$. We can make explicit the dependency on $T$. For instance, with cross-entropy we have:
\begin{align}
& \ell(\bm{e}^i, \hat{p}(\bm{y}|\bm{x})) = - \log~\hat{p}(\tilde{\bm{y}} = \bm{e}^i |\bm{x}) \\
& = - \log~\sum\nolimits_{j = 1}^{c}~p(\yCorrupt=\bm{e}^i | \bm{y}=\bm{e}^j)~\hat{p}(\bm{y} = \bm{e}^j |\bm{x})\\
& = - \log~\sum\nolimits_{j = 1}^{c}~T_{ji}~\hat{p}(\bm{y} = \bm{e}^j|\bm{x})\:\:,
\end{align}
\noindent or in matrix form
$
\bm{\ell}(\hat{p}(\bm{y}|\bm{x})) = - \log ~T^\top \hat{p}(\bm{y}|\bm{x})\:.
$ This loss compares the noisy label $\tilde{\bm{y}}$ to averaged noisy prediction corrupted by $T$. We call this procedure ``forward'' correction. In order to analyze its behavior, we first need to recall definition and properties of a broad family of losses named \emph{proper composite} \cite[Section 4]{rwCB}. Consider a \emph{link function} $\bm{\bm{\psi}}: \Delta^{c-1} \rightarrow \Real^c$, invertible. 
Many losses are said to be \emph{composite}, and denoted by $\ell_{\bm{\psi}}: \mathcal{Y} \times \Real^c \rightarrow \Real$, in the sense that they can be expressed by the aid of a link function as
\begin{align}
\ell_{\bm{\psi}}(\bm{y}, \bm{h}(\bm{x})) = \bm{\ell}(\bm{y}, \bm{\psi}^{-1}(\bm{h}(\bm{x})))\:\:.
\end{align}
\noindent In the case of cross-entropy, the softmax is the \emph{inverse} link function. When composite losses are also \emph{proper} \cite{rwCB}, their minimizer assumes the particular shape of the link function applied to the class-conditional probabilities $p(\bm{y}|\bm{x})$:
\begin{align}
\argmin_{\bm{h}} \mathbb{E}_{\bm{x}, \bm{y}}~\ell_{\bm{\psi}}(\bm{y}, \bm{h}(\bm{x})) = \bm{\psi}(p(\bm{y}|\bm{x}))\:\:.\label{eq:proper}
\end{align}
Cross-entropy and square are examples of \emph{proper composite} losses. An intriguing robustness property holds for forward correction of proper composite losses.

\begin{theorem}\label{th:forward}
Suppose that the noise matrix $T$ is non-singular. Given a proper composite loss $\bm{\ell}_{\bm{\psi}}$, define the \emph{forward} loss correction as:
$$\ellFb_{\bm{\psi}}(\bm{h}(\bm{x})) = \bm{\ell}(T^\top \bm{\psi}^{-1}(\bm{h}(\bm{x}))) \:\:.$$
Then, the minimizer of the corrected loss under the noisy distribution is the same as the minimizer of the original loss under the clean distribution:
\begin{align}
 \argmin_{\bm{h}}~\mathbb{E}_{\bm{x}, \tilde{\bm{y}}}~\ellFbScalar_{\bm{\psi}}(\bm{y}, \bm{h}(\bm{x})) 
  =\argmin_{\bm{h}}~\mathbb{E}_{\bm{x}, \bm{y}}~\ell_{\bm{\psi}}(\bm{y}, \bm{h}(\bm{x})) \nonumber
\:\:.
\end{align}
\end{theorem}
\textbf{Proof}. First notice that:
\begin{align}
\ellFbScalar_{\bm{\psi}}(\bm{y}, \bm{h}(\bm{x})) = \ell(\bm{y}, T^\top \bm{\psi}^{-1}(\bm{h}(\bm{x}))) = \ell_{\bm{\phi}}(\bm{y}, \bm{h}(\bm{x}))\label{proof:forward} \end{align}
\noindent where we denote $\bm{\phi}^{-1} = \bm{\psi}^{-1} \circ T^\top$. Equivalently, $\bm{\phi} =  (T^{-1})^\top \circ \bm{\psi}$ is invertible by composition of invertible functions, its domain is $\Delta^{c-1}$ as of $\bm{\psi}$ and its codomain is $\Real^c$. The last loss in Equation \ref{proof:forward} is therefore proper composite with link $\bm{\phi}$. Finally, from Equation \ref{eq:proper}, the loss minimizer over the noisy distribution is
\begin{align}
& \argmin_{\bm{h}} \mathbb{E}_{\bm{x}, \tilde{\bm{y}}}~\ell_{\bm{\phi}}(\bm{y}, \bm{h}(\bm{x})) = \bm{\phi}(p(\tilde{\bm{y}}|\bm{x})) \\
& = \bm{\psi}((T^{-1})^\top p(\tilde{\bm{y}}|\bm{x})) = \bm{\psi}(p(\bm{y}|\bm{x}))\:\:,
\end{align}
\noindent that proves the Theorem by Equation \ref{eq:proper} once again. \hfill $\blacksquare$

Recall that $\hat{p}(\bm{y}|\bm{x})$ approximates $p(\bm{y}|\bm{x})$ and thus we can relate to the result by taking any neural network that enough expressive.
Although, the property is weaker than unbiasedness of Theorem \ref{th:unbias}. Robustness applies to the minimizer only, that is, the model learned by forward correction is the minimizer over the \emph{clean} distribution. Yet, Theorem \ref{th:forward} guarantees noise robustness with no explicit matrix inversion; the ``de-noising'' link function $\bm{\phi}$ does it behind the scene. This turns out to be an important factor in practice; see below.

\subsection{The overall algorithm}
\label{sec:noise-estimation}

A limitation of the above procedures is that they require knowing $T$.
In most applications, the matrix $T$ would be unknown and to be estimated. We present here an extension of the recent noise estimator of \cite{ltCW, mvrowLF} to the multi-class settings. It is derived under two assumptions.

\begin{theorem}\label{th:noise-estimate}
Assume $p(\bm{x}, \bm{y})$ is such that:
\begin{enumerate}[(1)]
\item There exist 
``perfect examples'' of each of class $j \in [c]$, in the sense that
$$ ( \exists \bar{\bm{x}}^j \in \mathcal{ X } ) \, \colon~~p( \bar{\bm{x}}^j ) > 0 \land p(\bm{y}=\bm{e}^j | \bar{\bm{x}}^j ) = 1.$$
\item given sufficiently many corrupted samples, $\bm{h}$ is rich enough to model $p(\yCorrupt| \bm{x})$ accurately.
\end{enumerate}
It follows that
$\:\: \forall i, j \in [c],~~T_{i j} = p(\yCorrupt=\bm{e}^j | \bar{\bm{x}}^i)\:\:.$
\end{theorem}
\noindent \textbf{Proof}. By (2), we can consider $p(\tilde{\bm{y}}|\bm{x})$ instead of $\hat{p}(\tilde{\bm{y}}|\bm{x})$. For any $j \in [c]$ and any $\bm{x} \in \mathcal{X}$, we have that:
\begin{align}
p(\yCorrupt=\bm{e}^j | \bm{x}) & = \sum\nolimits_{k =1}^c p(\yCorrupt = \bm{e}^j | \bm{y} = \bm{e}^k)~p(\bm{y}=\bm{e}^k | \bm{x}) \nonumber \\
& = \sum\nolimits_{k =1}^c T_{kj}~p(\bm{y}=\bm{e}^k | \bm{x})\:\:.
\end{align}
\noindent By (1), when $\bm{x} = \bar{\bm{x}}^i$, $p(\bm{y}=\bm{e}^k | \bar{\bm{x}}^i) = 0$ for $k \neq i$. \hfill $\blacksquare$

Rather surprisingly, Theorem \ref{th:noise-estimate} tells us that we can estimate each component of matrix $T$ \emph{just based on noisy class probability estimates},
that is, the output of the softmax of a network trained with noisy labels.
In particular, let $X'$ be any set of features vectors. This can be the training set itself, but not necessarily: we do not require this sample to have \emph{any} label at all and therefore any unlabeled sample from the same distributions can be used as well.
We can approximate $T$ with two steps:
\begin{align}
\bar{\bm{x}}^i &= \argmax\nolimits_{\bm{x} \in X'}~\hat{p}(\tilde{\bm{y}} = \bm{e}^i|\bm{x}) \label{eq:est1} \\
\hat{T}_{ij} &= \hat{p}(\tilde{\bm{y}} = \bm{e}^j|\bar{\bm{x}}^i)\label{eq:est2}\:\:.
\end{align}

In practice, assumption (1) of Theorem \ref{th:noise-estimate} might hold true when $X'$ is large enough.
Assumption (2) of Theorem \ref{th:noise-estimate} is more difficult to justify; we require that the network can perfectly model the probability of \emph{the noisy labels}. Although, in the experiments we can often recover $T$ close to the ground truth and find that small estimation errors have a mild, not catastrophic effect on the quality of the correction.

\begin{algorithm}[t!]
\caption{Robust two-stage training}\label{algo:1}
\small
\begin{algorithmic}
\STATE \textbf{Input}: the noisy training set $\mathcal{S}$, any loss $\ell$
\STATE If $T$ is unknown:
\STATE \qquad Train a network $\bm{h}(\bm{x})$ on $\mathcal{S}$ with loss $\ell$
\STATE \qquad Obtain an unlabeled sample $X'$
\STATE \qquad Estimate $\hat{T}$ by Equations (\ref{eq:est1})-(\ref{eq:est2}) on $X'$
\STATE Train the network $\bm{h}(\bm{x})$ on $\mathcal{S}$ with loss $\ellRobustScalar$ or $\ellFbScalar$
\STATE \textbf{Output}: $\bm{h}(\cdot)$\;
\end{algorithmic}
\end{algorithm}

Algorithm \ref{algo:1} summarizes the end-to-end approach.
If we know $T$, for example by cleaning manually a subset of training data, we can train with $\ellRobustScalar$ or $\ellFbScalar$.
Otherwise, we first have to train the network with $\ell$ on noisy data,
and obtain from it estimates of $p(\yCorrupt | \bm{x})$ for each class via the output of the softmax.
After training $\hat{T}$ is computable in $O(c^2 \cdot |X'|)$. Finally, we re-train with the corrected loss, while potentially utilizing the first network to help initializing the second one.

\subsection{Digression: noise free Hessians via ReLU}

We now present a result of independent interest in the context of label noise.
The ReLU activation function appears to be a good fit
for an architecture in our noise model, since it brings the
particular convenience that the Hessian of the loss \textit{does not
depend on noise}, and hence the local curvature is left unchanged. At the same time, we are assured that backward correction by $T$ --- or any arbitrarily bad estimator of the matrix --- has no impact on those second order properties of the loss --- something that does not hold for the forward correction though. We stress the fact that other activation functions like the sigmoid do not share this guarantee. The proof makes use of the factorization trick due to \cite{pnncLF}.

\begin{theorem}\label{th:hessian}
Assume that all activation functions are ReLUs\footnote{A caveat: $\ell$ must be a linear-odd loss studied in \cite{pnncLF}; cross-entropy and square loss are such. At the same time, we could generalize Theorem \ref{th:hessian} to any neural network that expresses a piece-wise linear function, including for example max-pooling.}. Then, the Hessian of $\ell$ does not change under noise. Moreover, the Hessians of $\ellRobustScalar$ and $\ell$ are the same for any $T$.
\end{theorem}
\textbf{Proof}. We give the proof for cross-entropy for simplicity; see \cite{pnncLF} for a generalization.
When $\bm{y} = \bm{e}^i$ the loss is:
\begin{align*}
& - \log ~ \hat{p}(\bm{y} = \bm{e}^i|\bm{x})_i = - \log ~ \frac{e^{W^{(n)}_{i\cdot} \bm{x}^{(n-1)} + \bm{b}^{(n)}_i}}{\sum_{k=1}^c e^{W^{(n)}_{k \cdot} \bm{x}^{(n-1)} + \bm{b}^{(n)}_k}} \nonumber \\
& = - W^{(n)}_{i \cdot} \bm{x}^{(n-1)} + \bm{b}^{(n)}_i
 + \log ~ \sum\nolimits_{k=1}^c e^{W^{(n)}_{k \cdot} \bm{x}^{(n-1)} + \bm{b}^{(n)}_k}\:\:.
\end{align*}
\noindent The only dependence on the true class $\bm{e}^i$ above are the first two terms.
The log-partition is \emph{independent} of the precise class $i$.
Evidently, the noise affects the loss only through $W^{(n)}_{\cdot i}$ and $\bm{b}_i^{(n)}$: those are the \emph{only} terms in which $\ell(\bm{y}, \hat{p}(\bm{y}|\bm{x}))$ and $\ell(\yCorrupt, \hat{p}(\bm{y}|\bm{x}))$ may differ. Therefore we can rewrite the backward corrected loss as:
\begin{align}
& \ellRobustScalar(\bm{e}^j, \hat{p}(\bm{y}|\bm{x})) = ~\Big(T^{-1} \bm{\ell}(\hat{p}(\bm{y}|\bm{x})) \Big)_{j} \\
& =- \left(T^{-1} W^{(n)}\right)_{j\cdot} \bm{x}^{(n-1)} - \left(T^{-1} \bm{b}^{(n)}\right)_{j} \\
&~~~~ + \log ~ \sum\nolimits_{k=1}^c e^{W^{(n)}_{k \cdot} \bm{x}^{(n-1)} + \bm{b}^{(n)}_k}\:\:.
\end{align}
\noindent In fact, note that $T^{-1}$ does not affect the log-partition function. To see this, let $A(\bm{x}) = \log ~ (\sum_{k=1}^c e^{W^{(n)}_{k \cdot} \bm{x}^{(n-1)} + \bm{b}^{(n)}_k})$, with the (vector) log-partition being $A(\bm{x}) \bm{1}$. It follows that its correction is $T^{-1} A(\bm{x}) \bm{1} = A(\bm{x}) \bm{1}$, by left-multiplication of $T$ and because $T \bm{1} = \bm{1}$ since $T$ is row-stochastic. Thus
$
\ellRobustScalar(\bm{e}^j, \bm{h}_{s}\left(\bm{x})\right) = B(\bm{x} ) + A(\bm{x} ),
$
where $B(\bm{x} ) = - (T^{-1} W^{(n)})_{j\cdot} \bm{x}^{(n-1)} - (T^{-1} \bm{b}^{(n)})_j$ is a \textit{piece-wise linear} function of the \textit{model parameters}, and the log-partition $A(\bm{x})$ is non-linear because of the loss and the architecture but \textit{does not depend on noise}.
Since the composition of piece-wise linear function is piece-wise linear, the Hessian of $B(\bm{x} )$ vanishes, and therefore the Hessian of $\ellRobustScalar$ is noise independent for any $T$. The same holds for $\ell$ (no correction) by taking $T = I$ and hence the Hessians are the same. \hfill $\blacksquare$

\begin{table}[tb]
\footnotesize
\centering
\renewcommand{\arraystretch}{0.9}
\begin{tabular}{llll}
\toprule
loss & correction & $\mathbb{E}_{\bm{x}, \tilde{\bm{y}}}$ & Hessian of $\mathbb{E}_{\bm{x}, \tilde{\bm{y}}}$ \\
\midrule
$\ell$ & - & no guarantee  & unchanged \\
$\ellRobustScalar$  & $T^{-1} \cdot $ & unbiased estimator of $\ell$  & unchanged \\ 
$\ellFbScalar$ & $T \cdot $ & same minimizer of $\ell$ & no guarantee \\
\bottomrule
\end{tabular}
\caption{Qualitative comparison of loss corrections.}
\label{table:losses}
\end{table}

Theorem \ref{th:hessian} does not provide any assurance on minima: indeed, stationary points may change location due to label noise. What it \emph{does} guarantee is that the convergence rate of first-order methods is the same: the loss curvature cannot blow up or flat out and instead it is the same point by point in the model space. 
The Theorem advocates for use of ReLU networks, in line with the recent theoretical breakthrough allowing for deep learning with no local minima \cite{kDL}.
Table \ref{table:losses} summaries the properties of loss correction.

\section{Experiments}\label{sec:exp}

We now test the theory on various deep neural networks trained on MNIST \cite{lbbhGB}, IMDB \cite{mdphnpLW}, CIFAR-10, CIFAR-100 \cite{khLM} and Clothing1M \cite{xxtwLF} so as to stress that our approach is independent on both architecture and data domain.

\subsection{Loss corrections with $T$ known or estimated}\label{sec:exp1}

We artificially corrupt labels by a parametric matrix $T$. The rationale is to mimic some of the structure of real mistakes for similar classes, \emph{e.g.} \textsc{cat} $\rightarrow$ \textsc{dog}. Transitions are parameterized by $N \in [0, 1]$ such that ground truth and wrong class have probability respectively of $1-N, N$. An example of $T$ used for MNIST with $N=0.7$ is on the left:
\begin{equation}
\tiny
\setlength{\arraycolsep}{2.1pt}
 \begin{bmatrix}
    1 & 0 & 0 & 0 & 0 & 0 & 0 & 0 & 0 & 0\\
    0 & 1 & 0 & 0 & 0 & 0 & 0 & 0 & 0 & 0\\
    0 & 0 & .3 & 0 & 0 & 0 & 0 & .7 & 0 & 0\\
    0 & 0 & 0 & .3 & 0 & 0 & 0 & 0 & .7 & 0\\
    0 & 0 & 0 & 0 & 1 & 0 & 0 & 0 & 0 & 0\\
    0 & 0 & 0 & 0 & 0 & .3 & .7 & 0 & 0 & 0\\
    0 & 0 & 0 & 0 & 0 & .7 & .3 & 0 & 0 & 0\\
    0 & .7 & 0 & 0 & 0 & 0 & 0 & .3 & 0 & 0\\
    0 & 0 & 0 & 0 & 0 & 0 & 0 & 0 & 1 & 0\\
    0 & 0 & 0 & 0 & 0 & 0 & 0 & 0 & 0 & 1\\
\end{bmatrix},
\begin{bmatrix}
    1 & \epsilon & \epsilon & \epsilon & \epsilon & \epsilon & \epsilon & \epsilon & \epsilon & \epsilon\\
    \epsilon & 1 & \epsilon & \epsilon & \epsilon & \epsilon & \epsilon & \epsilon & \epsilon & \epsilon\\
    \epsilon & \epsilon & .33 & \epsilon & \epsilon & \epsilon & \epsilon & .67 & \epsilon & \epsilon\\
    \epsilon & \epsilon & \epsilon & .35 & \epsilon & \epsilon & \epsilon & \epsilon & .65 & \epsilon\\
    \epsilon & \epsilon & \epsilon & \epsilon & 1 & \epsilon & \epsilon & \epsilon & \epsilon & \epsilon\\
    \epsilon & \epsilon & \epsilon & \epsilon & \epsilon & .29 & .71 & \epsilon & \epsilon & \epsilon\\
    \epsilon & \epsilon & \epsilon & \epsilon & \epsilon & .73 & .26 & \epsilon & \epsilon & \epsilon\\
    \epsilon & .75 & \epsilon & \epsilon & \epsilon & \epsilon & \epsilon & .25 & \epsilon & \epsilon\\
    \epsilon & \epsilon & \epsilon & \epsilon & \epsilon & \epsilon & \epsilon & \epsilon & 1 & \epsilon\\
    \epsilon & \epsilon & \epsilon & \epsilon & \epsilon & \epsilon & \epsilon & \epsilon & \epsilon & 1\\
\end{bmatrix}\label{eq:bigT} 
\end{equation}

Common to all experiments is what follows. The loss $\ell$ chosen for comparison is cross-entropy. $10\%$ of training data is held out for validation. \emph{The loss} is evaluated on it during training. With the corrected losses we can validate \emph{on noisy data}, which is advantageous over other approaches that measure noisy validation accuracy instead. The available standard test sets are used for testing.
We use ReLU for all networks and initialize weights prior to ReLUs as in \cite{hzrsDD}, otherwise by uniform sampling in $[-0.05, 0.05]$.
The mini-batch size is $128$. The estimator of $T$ from noisy labels is applied to $X'$ being training and validation sets together. In fact, preliminary experiments highlighted that the large size $X'$ improve sensibly the approximation of $T$; after estimation, we row-normalize the matrix. Following \cite{mvrowLF}, we take a $\alpha$-percentile in place of the $\argmax$ of Equation \ref{eq:est1}, and we found $\alpha=97\%$ to work well for most experiments; the estimator performs very poorly with CIFAR-100, possibly due the small number of images per class, and we found it is better off computing the $\argmax$ instead.

\begin{figure}[t]
\centering
\begin{subfigure}[b]{0.23\textwidth}
\includegraphics[width=\textwidth]{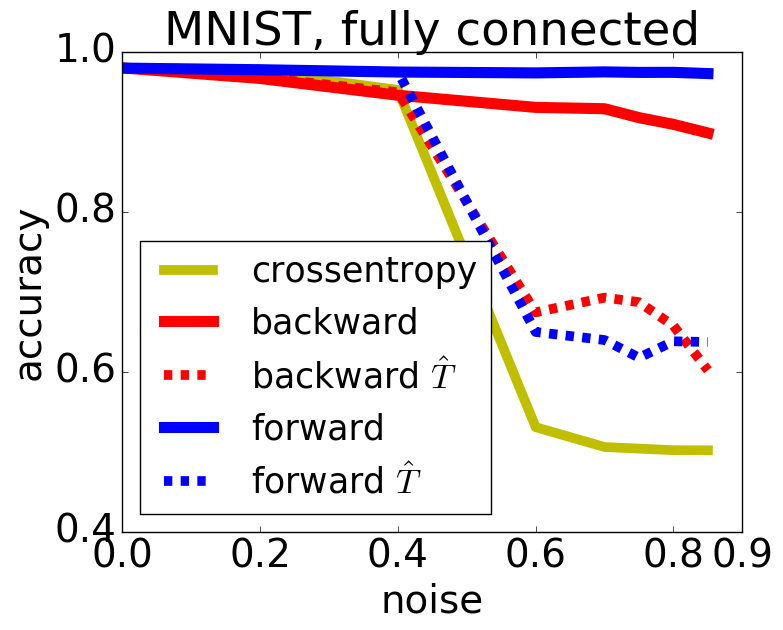}\caption{} \label{fig:mnist}
 \includegraphics[width=\textwidth]{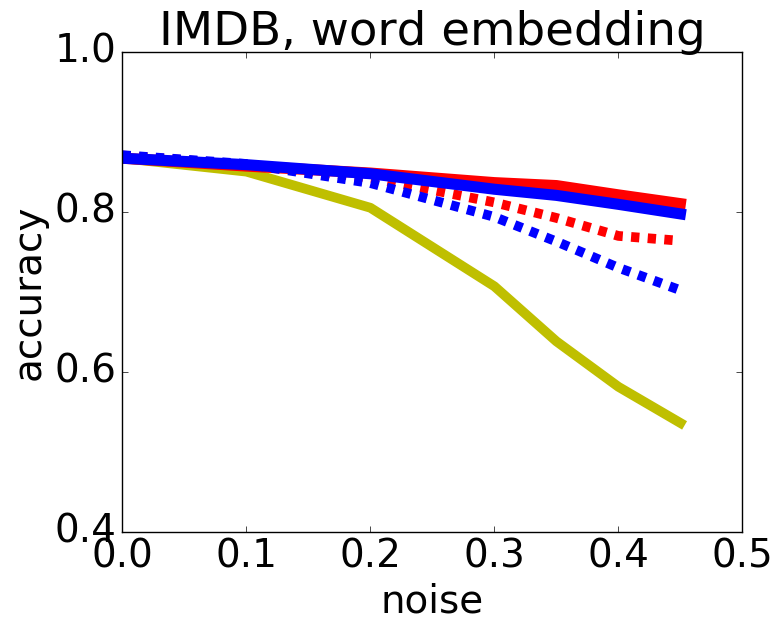}\caption{} \label{fig:imdb}
 \includegraphics[width=\textwidth]{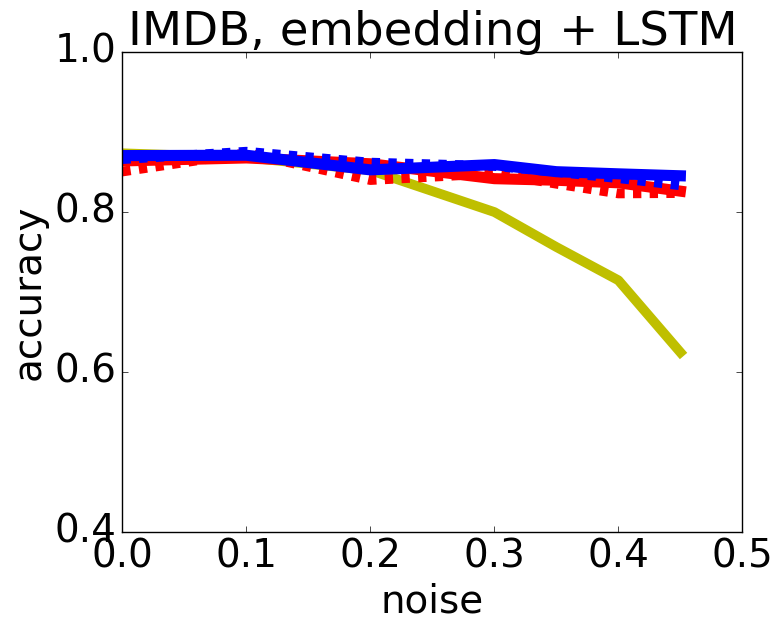}\caption{} \label{fig:lstm} 
\end{subfigure} 
\begin{subfigure}[b]{0.23\textwidth}
\includegraphics[width=\textwidth]{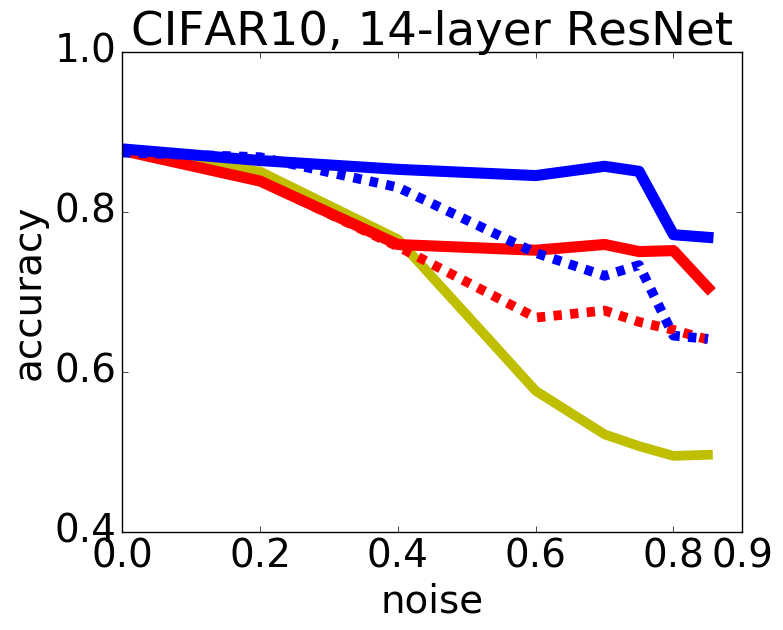}\caption{} \label{fig:cifar10_resnet1} 
\includegraphics[width=\textwidth]{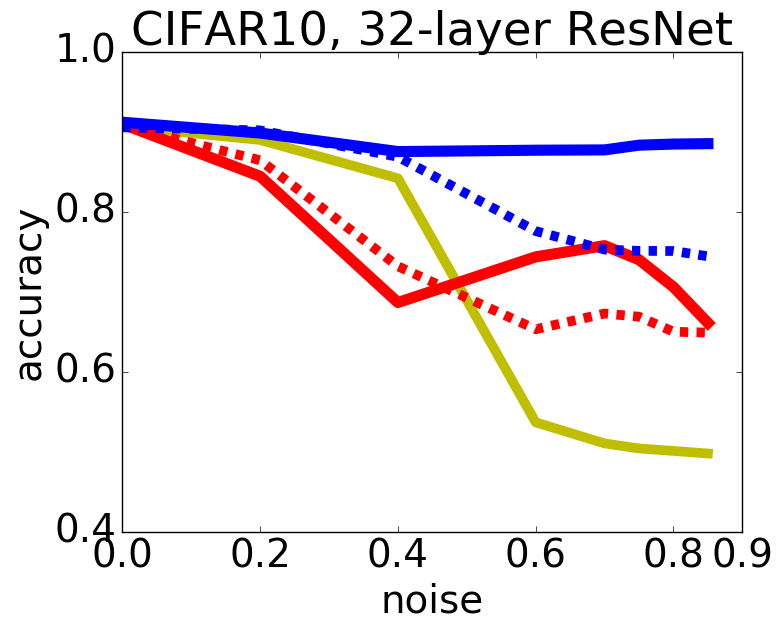}\caption{} \label{fig:cifar10_resnet5}
\includegraphics[width=\textwidth]{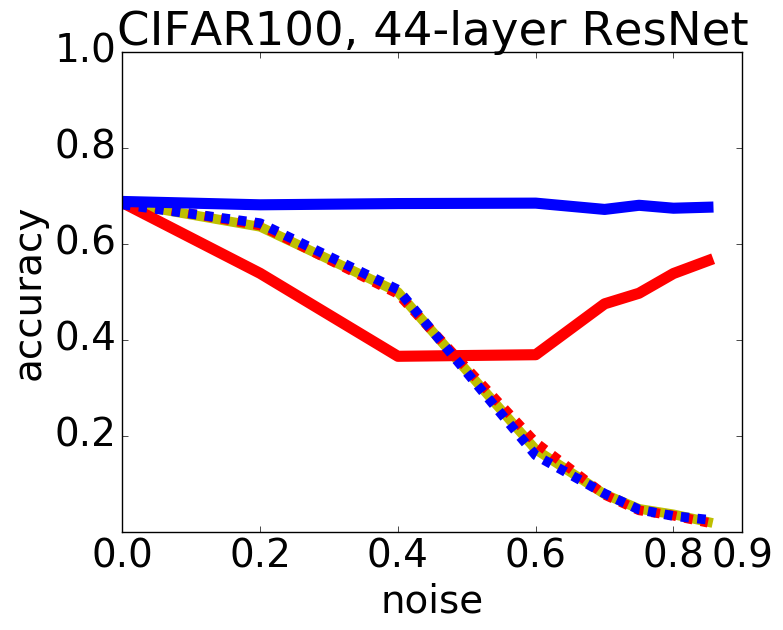}\caption{} \label{fig:cifar100}
\end{subfigure}
\caption{Comparison of cross-entropy with its corrections, with known or estimated $T$.}
\vspace{-0.3cm}
\end{figure}

\textit{Fully connected network on MNIST.} In the first set of experiments we consider MNIST. Pixels are normalized in $[0,1]$. Noise flips some of the similar digits: $2 \rightarrow 7, 3 \rightarrow 8, 5 \leftrightarrow 6, 7 \rightarrow 1$; see Equation (\ref{eq:bigT}, left). We train an architecture with two dense hidden layers of size $128$, with probability $0.5$ of dropout. AdaGrad \cite{asDHS} is run for 40 epochs with initial learning rate $0.01$ and $\delta=10^{-6}$. We repeat each experiment 5 times to account for noise and weight initialization. It is clear from Figure \ref{fig:mnist} that, although the model is somewhat robust to mild noise, high level of corruption has a disrupting effect on $\ell$. Instead, our losses do not witness a drastic drop. With $\hat{T}$ estimated performance lays in between, yet it is significantly better than with no correction. An example of $\hat{T}$ is in Equation (\ref{eq:bigT}, right), with $\epsilon < 10^{-6}$.

\textit{Word embedding and LSTM on IMDB.}
We keep only the top 5000 most frequent words in the corpus. Each review is either truncated or padded to be 400-word long. To simulate asymmetric noise in this binary problem, we keep constant noise for the transition $0 \rightarrow 1$ at $5\%$, while $1 \rightarrow 0$  is parameterized as above; $0/1$ are the two review's sentiments. We trained two models inspired by the baselines of \cite{dlSS}. The first maps words into $50$-dimensional embeddings, before passing through ReLUs; dropout with probability $0.8$ is applied to the embedding output. In the second model the embedding has dimension $256$ and it is followed by an LSTM with $512$ units and by a last $512$-dimensional hidden layer with $0.5$ dropout. AdaGrad is run for 50 epochs with the same setup as above; results are averages over 5 runs. Figures \ref{fig:imdb}-\ref{fig:lstm} display an outcome similar to what previously observed on MNIST, in spite of difference in dataset, number of classes, architecture and structure of $T$. Noticeably, our approach is effective on recurrent networks as well. Correcting with $\hat{T}$ is in line with the true $T$ here; we believe this is because estimation is easier on this binary problem.

\textit{Residual networks on CIFAR-10 and CIFAR-100.} For both datasets we perform per-pixel mean subtraction and data augmentation as in \cite{He_2015}, by horizontal random flips and $32 \times 32$ random crops after padding with 4 pixels on each side. $T$ for CIFAR-10 is described by: \textsc{truck} $\rightarrow$ \textsc{automobile}, \textsc{bird} $\rightarrow$ \textsc{airplane}, \textsc{deer} $\rightarrow$ \textsc{horse}, \textsc{cat} $\leftrightarrow$ \textsc{dog}. In CIFAR-100, the 100 classes are grouped into $20$ $5$-size super-classes, \emph{e.g.} \textsc{aquatic} mammals contain \textsc{beaver, dolphin, otter, seal} and \textsc{whale}. Within super-classes, the noise flips each class into the next, circularly.

For the last experiments we use deep residual networks (ResNet), the CIFAR-10/100 architectures from \cite{He_2015}. In short, residual blocks implements a non-linear operation $F(\bm{x})$ in parallel with an identity shortcut: $\bm{x} \rightarrow \bm{x} + F(\bm{x})$.
$F$ is as cascade of twice batch normalization $\rightarrow$ ReLU $\rightarrow$ $3 \times 3$ convolution, following the ``pre-activation" recommendation of \cite{hzrsIM}. 
Here we experiment with ResNets of depth $14$ and $32$ (CIFAR-10) and $44$ (CIFAR-100). By common practice \cite{hslswDN}, we run SGD with $0.9$ momentum and learning rate $0.01$, and divide it by $10$ after $40$ and $80$ epoch ($120$ in total) for CIFAR-10 and after $80$ and $120$ ($150$) for CIFAR-100; weight decay is $10^{-4}$. Training deep ResNets is more time consuming and thus experiments are run only once. Since we use shallower networks than the ones in~\cite{He_2015}, performance is not comparable with the original work. In figures \ref{fig:cifar10_resnet1}-\ref{fig:cifar100}, forward correction does not suffer any significant loss. Except with the shallowest ResNet, backward correction does not seem to work well in the low noise regime. Finally, noise estimation is particularly difficult on CIFAR-100.

\subsection{Comparing with other loss functions}

\begin{table*}[tb]
\tiny
\centering
\begin{tabular}{l|cccc|ccccc}
\hline
\: & \multicolumn{4}{c}{MNIST, fully connected} & \multicolumn{4}{|c}{CIFAR-10, 14-layer ResNet} \\

 & \textsc{no noise}& \textsc{symm.} $N=0.2$ & \textsc{asymm.}  $N=0.2$ & \textsc{asymm.} $N=0.6$ &
 \textsc{no noise}& \textsc{symm.} $N=0.2$ & \textsc{asymm.} $N=0.2$ & \textsc{asymm.} $N=0.6$\\
 \hline
\hline
 cross-entropy &      $97.9 \pm 0.0$ & $\bm{96.9 \pm 0.1}$ & $97.5 \pm 0.0$ & $53.0 \pm 0.6$ & 
 $87.8$ & $83.7$ & $85.0$ & $57.6$  \\
 \hline
        unhinged (BN) &     $97.6 \pm 0.0$ & $\bm{96.9 \pm 0.1}$  & $97.0 \pm 0.1$ & $\bm{71.2 \pm 1.0}$  &
       $86.9$ & $\bm{84.1}$   & $83.8$ & $52.1$ \\
       sigmoid (BN) &      $97.2 \pm 0.1$ & $93.1 \pm 0.2$ &  $96.7 \pm 0.1$ & $\bm{71.4 \pm 1.3}$ &
       $76.0$ & $66.6$ &  $71.8$ & $57.0$\\
       Savage  &      $97.3 \pm 0.0$ & $\bm{96.9 \pm 0.0}$ & $97.0 \pm 0.1$ & $51.3 \pm 0.4$  &
       $80.1$ & $77.4$ & $76.0$ & $50.5$  \\
       bootstrap soft  &      $97.9 \pm 0.0$ & $\bm{96.9 \pm 0.0}$ & $97.5 \pm 0.0$ & $53.0 \pm 0.4$  &
       $87.7$ & $\bm{84.3}$  & $84.6$ & $57.8$  \\
       bootstrap hard  &      $97.9 \pm 0.0$ & $96.8 \pm 0.0$  & $97.4 \pm 0.0$ & $55.0 \pm 1.3$  &
       $87.3$ & $83.6$ & $84.7$ & $58.3$  \\
\hline
       backward  &      $97.9 \pm 0.0$ & $96.3 \pm 0.1$ & $96.6 \pm 1.1$ & $\:\: \bm{93.0 \pm 0.9}^\star$  &
       $87.6$ & $81.5$ &  $83.8$ & $\:\: \bm{75.2}^\star$  \\
       backward $\hat{T}$  &      $97.9 \pm 0.0$ & $\bm{96.9 \pm 0.0}$ & $96.7 \pm 0.1$ & $67.4 \pm 1.5$  &
       $87.7$ & $80.4$ & $83.8$ & $66.7$\\
       forward  &      $97.9 \pm 0.0$ & $\:\: \bm{97.3 \pm 0.0}^\star$ &  $\bm{97.7 \pm 0.0}$ & $\:\: \bm{97.3 \pm 0.0}^\star$  &
       $87.8$ & $\:\: \bm{85.6}^\star$ & $86.3$ & $\:\: \bm{84.5}^\star$  \\
       forward $\hat{T}$  &      $97.9 \pm 0.0$ & $\bm{96.9 \pm 0.0}$  & $\bm{97.7 \pm 0.0}$ & $64.9 \pm 4.4$  &
       $87.4$  & $83.4$ & $\bm{87.0}$ & $\bm{74.8}$ \\
 \hline
\hline
\: & \multicolumn{4}{c}{IMBD, word embedding} & \multicolumn{4}{|c}{CIFAR-10, 32-layer ResNet} \\

 & \textsc{no noise}& \textsc{symm.} $N=0.1$ & \textsc{asymm.} $N=0.1$& \textsc{asymm.} $N=0.4$ &
 \textsc{no noise}& \textsc{symm.} $N=0.2$ & \textsc{asymm.} $N=0.2$ & \textsc{asymm.} $N=0.6$\\
\hline 
\hline
 cross-entropy &      $86.7 \pm 0.0$ & $84.6 \pm 0.1$ & $85.0 \pm 0.2$ & $58.1 \pm 0.5$ &
 $90.1$ & $86.6$ & $89.0$ & $53.6$ \\
 \hline
        unhinged (BN) &     $83.3 \pm 0.0$ & $76.9 \pm 0.5$ & $80.6 \pm 0.3$ & $72.9 \pm 0.4$  &
        $90.2$ & $86.5$ & $87.1$ & $60.0$\\
       sigmoid (BN)  &      $84.3 \pm 0.0$ & $80.2 \pm 0.3$ & $81.7 \pm 0.5$ & $72.8 \pm 0.6$  &
        $81.6$ & $69.6$ & $79.1$ & $61.8$ \\
       Savage  &      $86.5 \pm 0.0$ & $84.3 \pm 0.4$ & $85.2 \pm 0.3$ & $58.3 \pm 1.0$  &
       $88.3$ & $86.2$ & $86.3$ & $53.5$\\
       bootstrap soft  &      $86.7 \pm 0.0$ & $84.5 \pm 0.1$ & $85.1 \pm 0.1$ & $57.8 \pm 0.7$  &
       $90.9$ & $86.9$ & $88.6$ & $53.1$ \\
       bootstrap hard  &      $86.7 \pm 0.0$ & $84.6 \pm 0.3$ & $85.1 \pm 0.3$ & $59.0 \pm 0.6$  &
       $90.4$ & $86.4$ & $88.6$  & $54.7$\\
\hline
       backward  &      $86.7 \pm 0.0$ & $\:\: \bm{85.3 \pm 0.3}^\star$ & $\bm{85.7 \pm 0.1}$ & $\:\: \bm{82.1 \pm 0.1}^\star$  &
       $90.1$ & $83.0$ & $84.4$ & $74.3$  \\
       backward $\hat{T}$  &      $\:\: 87.0 \pm 0.0^{\dagger}$ & $\bm{85.1 \pm 0.4}$ &  $\bm{85.8 \pm 0.2}$ & $\bm{77.0 \pm 1.4}$  &
       $90.8$ & $86.9$ & $86.4$ & $66.7$ \\
       forward  &      $86.7 \pm 0.0$ & $\:\: \bm{85.3 \pm 0.2}^\star$ & $\bm{85.9 \pm 0.1}$ & $\:\: \bm{80.9 \pm 1.3}^\star$  &
       $91.2$ & $\bm{87.7}$ & $\bm{89.9}$ & $\:\: \bm{87.6}^\star$ \\
       forward $\hat{T}$   &      $\:\: 87.0 \pm 0.0^{\dagger}$ & $\bm{85.2 \pm 0.3}$ &  $\bm{85.9 \pm 0.2}$ & $73.0 \pm 1.2$ &
        $90.5$ & $\bm{87.9}$ &  $\bm{90.1}$ & $\bm{77.6}$ \\
 \hline
\hline

\: & \multicolumn{4}{c}{IMBD, word embedding + LSTM} & \multicolumn{4}{|c}{CIFAR-100, 44-layer ResNet} \\

 & \textsc{no noise}& \textsc{symm.} $N=0.1$ & \textsc{asymm.} $N=0.1$& \textsc{asymm.} $N=0.4$ &
 \textsc{no noise}& \textsc{symm.} $N=0.2$ & \textsc{asymm.} $N=0.2$ & \textsc{asymm.} $N=0.6$\\
\hline 
\hline
 cross-entropy &      $87.8 \pm 0.4$ & $\bm{85.2 \pm 0.5}$&  $86.8 \pm 0.4$ & $71.4 \pm 1.3$ 
 & $68.5$ & $57.9$ & $63.5$ & $17.1$ \\
 \hline
        unhinged (BN) &     $84.3 \pm 4.4$ & $\:\: 69.7 \pm 15.9$ &  $85.2 \pm 1.2$ & $\:\: 59.4 \pm 12.9$  
       & $50.9$ & $47.5$ & $48.0$ & $14.5$\\
       sigmoid (BN) & $87.7 \pm 0.5$ & $\:\: 77.6 \pm 13.6$ & $\bm{86.3 \pm 3.1}$ & $\:\: \bm{70.0 \pm 14.6}$
       & $58.2$ & $47.6$ & $55.6$ & $16.4$ \\
       Savage  &      $87.4 \pm 0.3$ & $\bm{85.1 \pm 0.6}$ & $\bm{87.2 \pm 0.3}$ & $70.4 \pm 3.8$  
       & $1.4$ & $2.0$ & $1.8$ & $1.6$\\
       bootstrap soft  &      $87.1 \pm 0.6$ & $\bm{83.5 \pm 2.5}$ & $\bm{86.1 \pm 1.2}$ & $69.0 \pm 5.3$  
       & $67.9$ & $57.8$ & $\bm{63.8}$ & $16.3$ \\
       bootstrap hard  &      $86.5 \pm 0.5$ & $\bm{84.3 \pm 1.0}$ & $86.7 \pm 0.4$ & $71.8 \pm 3.3$  
       & $68.5$ & $57.3$ & $\bm{63.9}$& $17.0$\\
\hline
       backward  &      $87.6 \pm 0.2$ & $\bm{84.3 \pm 0.9}$ & $86.7 \pm 0.5$ & $\bm{83.6 \pm 1.4}$  
        & $68.5$ & $55.1$ & $53.8$ & $\:\: \bm{36.8}^\star$ \\
       backward $\hat{T}$  & $87.2 \pm 0.7$&  $\bm{82.8 \pm 2.7}$  & $\bm{87.3 \pm 0.1}$ & $\bm{82.3 \pm 1.7}$
         & $68.6$ & $51.7$ & $63.8$ & $\bm{18.5}$ \\
       forward  &      $87.5 \pm 0.2$ & $\bm{85.0 \pm 0.2}$ & $87.0 \pm 0.4$ & $\:\: \bm{84.7 \pm 0.6}^\star$  
        & $68.8$ & $\:\: \bm{64.0}^\star$ & $\:\: \bm{68.1}^\star$ & $\:\: \bm{68.4}^\star$ \\
       forward $\hat{T}$   & $\:\: 87.8 \pm 1.5^{\dagger}$  & $\bm{84.1 \pm 1.0}$ & $\bm{87.5 \pm 0.2}$ & $\bm{84.2 \pm 0.9}$
       & $68.1$ & $\bm{58.6}$ & $\bm{64.2}$ & $15.9$ \\
 \hline
\end{tabular}
\caption{Average accuracy with standard deviation (5 runs, left part) is bold faced when statistically far from the others, by means of passing a \emph{Welch's t-test} with $p$-value $<5\%$; in case the highest accuracy is due to $\ellRobustScalar$ or $\ellFbScalar$ with the ground truth $T$, we denote those by $^\star$ and highlight the next highest accuracy as well. For experiments with no standard deviation (right part), the same rule is applied, but bold face is given to the all accuracies in a range of $0.5$ points from the highest. The meaning of $N$ depends on symmetric \emph{vs.} asymmetric noise and on number of classes (see Section \ref{sec:exp1}). On the first columns with no injected noise, $\dagger$ indicates when the noise estimation recovers some \emph{natural} noise and beats ``loss correction'' with $T=I$.}
\label{table:big}
\vspace{-0.30cm}
\end{table*}

We now compare with other methods. Data, architectures and artificial noise are the same as above. Additionally, we test the case of symmetric noise where $N$ is the probability of label flip that is spread uniformly among all the other classes. We select methods prescribing changes in the loss function, similarly to ours: unhinged \cite{rmwLW}, sigmoid \cite{gmsMR}, Savage \cite{msvOT} and soft and hard bootstrapping \cite{rlazerTD}; hyper-parameters of the last two methods are set in accordance with their paper.

Unhinged loss is unbounded and cannot be used alone. In the original work $L_2$ regularization is applied to address the problem, when training non-parametric kernel models. We tried to regularize every layer with little success; learning either does not converge (too little regularization) or converge to very poor solutions (too much). On preliminary experiments sigmoid loss ran into the opposite issue, namely premature saturation; the loss reaches a plateau too quickly, a well-known problem with sigmoidal activation functions \cite{gbUT}. To make those losses usable for comparison, we stack a layer of batch normalization right before the loss function. Essentially, the network outputs are whitened and likely to operate in a bounded, non-saturated area of the loss; note that this is never required for linear or kernel models.

Table \ref{table:big} presents the empirical analysis. We list the key findings: (a) In the absence of artificial noise (first column for each dataset), all losses reach similar accuracies with a spread of $2$ points; exceptions are some instances of unhinged, sigmoid and Savage. Additionally, with IMDB there are cases ($\dagger$ in Table \ref{table:big}) of loss correction with noise estimation that perform slightly better than assuming no noise. Clearly, the estimator is able to recover the \emph{natural} noise in the sentiment reviews. (b) With low asymmetric noise (second column) results differ between simple architecture/tasks (datasets on the left) and deep networks/more difficult problems (right); in the former case, the two corrections behave similarly and are not statistically far from the competitors; in the latter case, forward correction with known $T$ is unbeaten, with no clear winner among the remaining ones. (c) With asymmetric noise (last two columns) the two loss corrections with known $T$ are overall the best performing, confirming the practical implications of their formal guarantees; forward is usually the best. (d) If we exclude CIFAR-100, the noise estimation accounts for average accuracy drops between $0$ (IMBD with LSTM model) and $27$ points (MNIST); nevertheless, our performance is better than every other method in many occasions. (e) In the experiment on CIFAR-100 we obtain essentially perfect noise robustness with the ideal forward correction. The noise estimation works well except in the very last column, yet it guarantees again better accuracy over competing methods. We discuss this issue in Section \ref{sec:disc}.

\subsection{Experiments on Clothing1M}

\begin{table}[tb]
\footnotesize
\centering
\renewcommand{\arraystretch}{0.8}
\begin{tabular}{l|l|c|c|c|c}
\hline
\multicolumn{6}{c}{Clothing1M} \\
\hline
$\#$ & model & loss & init & training & accuracy \\
\hline
	1 & AlexNet & cross-. & ImageNet  & $50k$ & $72.63$ \\
	2 & AlexNet \cite{sbpbfTC} & cross-. & $\# 1$ & $1M, 50k$  & $76.22$ \\
	3 & AlexNet \cite{xxtwLF} & cross-. & $\# 1$ & $1M, 50k$ & $78.24$ \\
 \hline
   	4 & 50-ResNet & cross- & ImageNet  & $1M$ & $68.94$ \\
  	5 & 50-ResNet & backward & ImageNet  & $1M$  & $69.13$ \\
  	6 & 50-ResNet & forward & ImageNet  & $1M$  & $69.84$ \\
 	7 & 50-ResNet & cross-. &  ImageNet  & $50k$  & $75.19$ \\
 	8 & 50-ResNet & cross-. & $\# 6$ & $50k$ & $\bm{80.38}$ \\
\hline
\hline
\end{tabular} 
\caption{Results on the top section are from \cite{xxtwLF}. In $\# 2, \# 3$ the clean $50k$ are bootstrapped to $500k$. Best result $\# 8$ is obtained by fine tuning a net trained with forward correction.}
\label{table:last}
\vspace{-0.35cm}
\end{table}

Finally, we test on Clothing1M \cite{xxtwLF}, consisting of 1M images with noisy labels, with additional $50k, 14k, 10k$ of clean data respectively for training, validation and testing; we refer to those sets by their size. We aim to classify images within 14 classes, \emph{e.g.} t-shirt, suit, vest. In the original work two AlexNets \cite{kshIC} are trained together via EM; the networks are pre-trained with ImageNet. Two practical tricks are fundamental: a first learning phase with the clean $50k$ to help EM ($\# 1$ in Table \ref{table:last}) and a second phase with the mix of $50k$ bootstrapped to $500k$ and $1M$ ($\# 3$). Data augmentation is also applied, same as in Section \ref{sec:exp1} for CIFAR-10.

We learn a 50-layer ResNet pre-trained on ImageNet --- the bottleneck architecture of \cite{He_2015} --- with SGD with learning rate $10^{-3}$ and $10^{-4}$ for 5 epochs each, $0.9$ momentum, and batch size $32$. When we train with $50k$ we use weight decay of $5 \cdot 10^{-2}$ and data augmentation, while with $1M$ we use only weight decay of $10^{-3}$. The ResNet gives an uplift of about $2.5\%$ by training with $50k$ only ($\# 7$ \emph{vs.} $\# 1$). However, the large amount of noisy images is essential to compete with $\# 3$. Instead of estimating the matrix $T$ by (\ref{eq:est1})-(\ref{eq:est2}), we exploit the curated labels of $50k$ and their noisy versions in $1M$.  Forward and backward corrections are confirmed to work better than cross-entropy ($\# 6, \# 5$ \emph{vs.} $\# 4$), yet cannot reach the state of the art without the additional clean data. Thus, we fine tune the networks with $50k$, with the same learning parameters as in $\# 7$; due to reasons of time we only tune $\# 6$. The new state of the art is $\# 8$ that outperforms \cite{xxtwLF} of more than $2$ percent, which is achieved without time consuming bootstrapping of the $50k$.


\section{Discussion and Conclusion}\label{sec:disc}

We have proposed a framework for training deep neural networks with noisy labels that boils down to two loss corrections. Accuracy is consistently only few percent points away from training cross-entropy \emph{on clean data}, while corruption can worsen performance of cross-entropy by $40$ percent or more. Forward correction often performs better.
We believe the reason is not statistical --- Theorems \ref{th:unbias} and \ref{th:forward} guarantee optimality, in the limit of infinite data. The cause may be either numerical (via matrix inversion) or a drastic change of the loss (in particular its Hessian), which may have a detrimental effect on optimization. Indeed, backward correction is a linear combination of losses for every possible label, with coefficients that can be far by orders of magnitude and thus makes the learning harder. Instead, forward correction projects predictions into a probability distribution in $[0,1]$. 

The quality of noise estimation is a key factor for obtaining robustness. In practice, it works well in most experiments with a median drop of only $10$ points of accuracy with respect to using the true $T$. The exception is the last column for CIFAR-100, where estimation destroys most of the gain from loss correction. We believe that the mix of high noise and limited number of images per class (500) is detrimental to the estimator. This is confirmed by the sensitivity of $\alpha$.


Future work shall improve the estimation phase by incorporating priors of the noise structure, for example assuming low rank $T$. Improvements on this direction may also widen the applicability to massively multi-class scenarios. It remains an open question whether \emph{instance}-dependent noise may be included into our approach \cite{xxtwLF, mrnBL}. Finally, we anticipate the use of our approach as a tool for pre-training models with noisy data from the Web, in the spirit of \cite{kshtdpfTU}. \\

\noindent \textbf{Acknowledgments.} \textit{We gratefully acknowledge NVIDIA Corporation for the donation of a Tesla K40 GPU. We also thank the developers of Keras \cite{keras}.}

{\small
\bibliographystyle{ieee}
\bibliography{bib}

\begin{thebibliography}{10}\itemsep=-1pt

\bibitem{keras}
F.~Chollet.
\newblock {Keras}.
\newblock \url{github.com/fchollet/keras}.

\bibitem{dlSS}
A.~M. Dai and Q.~V. Le.
\newblock Semi-supervised sequence learning.
\newblock In {\em NIPS*29}, 2015.

\bibitem{dfgLE}
S.~Divvala, A.~Farhadi, and C.~Guestrin.
\newblock Learning everything about anything: Webly-supervised visual concept
  learning.
\newblock In {\em 27$^{th}$ IEEE CVPR}, 2014.

\bibitem{asDHS}
J.~Duchi, E.~Hazan, and Y.~Singer.
\newblock Adaptive subgradient methods for online learning and stochastic
  optimization.
\newblock {\em JMLR}, 12:2121--2159, 2011.

\bibitem{ffpzLO}
R.~Fergus, L.~Fei-Fei, P.~Perona, and A.~Zisserman.
\newblock Learning object categories from internet image searches.
\newblock {\em Proceedings of the IEEE}, 98(8):1453--1466, 2010.

\bibitem{fkesbhEA}
M.~Feurer, A.~Klein, K.~Eggensperger, J.~Springenberg, M.~Blum, and F.~Hutter.
\newblock Efficient and robust automated machine learning.
\newblock In {\em NIPS*29}, 2015.

\bibitem{Frenay_2014}
B.~Fr\'{e}nay and M.~Verleysen.
\newblock {Classification in the Presence of Label Noise: A Survey}.
\newblock {\em IEEE Transactions on Neural Networks and Learning Systems},
  25(5):845--869, May 2014.

\bibitem{gmsMR}
A.~Ghosh, N.~Manwani, and P.~S. Sastry.
\newblock Making risk minimization tolerant to label noise.
\newblock {\em Neurocomputing}, 2015.

\bibitem{gbUT}
X.~Glorot and Y.~Bengio.
\newblock Understanding the difficulty of training deep feedforward neural
  networks.
\newblock In {\em AISTATS}, 2010.

\bibitem{hzrsDD}
K.~He, X.~Zhang, S.~Ren, and J.~Sun.
\newblock Delving deep into rectifiers: Surpassing human-level performance on
  imagenet classification.
\newblock In {\em ICCV}, 2015.

\bibitem{He_2015}
K.~He, X.~Zhang., S.~Ren, and J.~Sun.
\newblock Deep residual learning for image recognition.
\newblock In {\em 29$^{th}$ IEEE CVPR}, 2016.

\bibitem{hzrsIM}
K.~He, X.~Zhang, S.~Ren, and J.~Sun.
\newblock Identity mappings in deep residual networks.
\newblock In {\em 14$^{~th}$ ECCV}, 2016.

\bibitem{hsLS}
S.~Hochreiter and J.~Schmidhuber.
\newblock Long short-term memory.
\newblock {\em Neural computation}, 9(8):1735--1780, 1997.

\bibitem{hslswDN}
G.~Huang, Y.~Sun, Z.~Liu, D.~Sedra, and K.~Weinberger.
\newblock Deep networks with stochastic depth.
\newblock In {\em 14$^{~th}$ ECCV}, 2016.

\bibitem{isBN}
S.~Ioffe and C.~Szegedy.
\newblock Batch normalization: Accelerating deep network training by reducing
  internal covariate shift.
\newblock In {\em 32$^{~th}$~ICML}, 2015.

\bibitem{kDL}
K.~Kawaguchi.
\newblock Deep learning without poor local minima.
\newblock In {\em NIPS*30}, 2016.

\bibitem{kshtdpfTU}
J.~Krause, B.~Sapp, A.~Howard, H.~Zhou, A.~Toshev, T.~Duerig, J.~Philbin, and
  L.~Fei-Fei.
\newblock The unreasonable effectiveness of noisy data for fine-grained
  recognition.
\newblock In {\em 14$^{~th}$ ECCV}, 2016.

\bibitem{khLM}
A.~Krizhevsky and G.~Hinton.
\newblock Learning multiple layers of features from tiny images.
\newblock Technical report, University of Toronto, 2009.

\bibitem{kshIC}
A.~Krizhevsky, I.~Sutskever, and G.~E. Hinton.
\newblock Imagenet classification with deep convolutional neural networks.
\newblock In {\em NIPS*26}, 2012.

\bibitem{lbbhGB}
Y.~LeCun, L.~Bottou, Y.~Bengio, and P.~Haffner.
\newblock Gradient-based learning applied to document recognition.
\newblock {\em Proceedings of the IEEE}, 86(11):2278--2324, 1998.

\bibitem{ltCW}
T.~Liu and D.~Tao.
\newblock Classification with noisy labels by importance reweighting.
\newblock {\em IEEE Transactions on PAMI}, 38(3):447--461, 2016.

\bibitem{lsRC}
P.~M. Long and R.~A. Servedio.
\newblock Random classification noise defeats all convex potential boosters.
\newblock {\em Machine learning}, 78(3):287--304, 2010.

\bibitem{mdphnpLW}
A.~L. Maas, R.~E. Daly, P.~T. Pham, D.~Huang, A.~Y. Ng, and C.~Potts.
\newblock Learning word vectors for sentiment analysis.
\newblock In {\em 49$^{~th}$ ACL}, 2011.

\bibitem{msvOT}
H.~Masnadi-Shirazi and N.~Vasconcelos.
\newblock On the design of loss functions for classification: theory,
  robustness to outliers, and savageboost.
\newblock In {\em NIPS*23}, 2009.

\bibitem{mrnBL}
A.~Menon, B.~{van Rooyen}, and N.~Natarajan.
\newblock Learning from binary labels with instance-dependent corruption.
\newblock {\em arXiv preprint arXiv:1605.00751}, 2016.

\bibitem{mvrowLF}
A.~Menon, B.~{van Rooyen}, C.~S. Ong, and B.~Williamson.
\newblock Learning from corrupted binary labels via class-probability
  estimation.
\newblock In {\em 32$^{~th}$~ICML}, 2015.

\bibitem{mhLT}
V.~Mnih and G.~E. Hinton.
\newblock Learning to label aerial images from noisy data.
\newblock In {\em 29$^{~th}$~ICML}, 2012.

\bibitem{ndrtLW}
N.~Natarajan, I.~S. Dhillon, P.~K. Ravikumar, and A.~Tewari.
\newblock Learning with noisy labels.
\newblock In {\em NIPS*27}, 2013.

\bibitem{nlxVR}
L.~Niu, W.~Li, and D.~Xu.
\newblock Visual recognition by learning from web data: A weakly supervised
  domain generalization approach.
\newblock In {\em 28$^{th}$ IEEE CVPR}, 2015.

\bibitem{pnncLF}
G.~Patrini, F.~Nielsen, R.~Nock, and M.~Carioni.
\newblock Loss factorization, weakly supervised learning and label noise
  robustness.
\newblock In {\em 33$^{~th}$~ICML}, 2016.

\bibitem{Ramaswamy:2016}
H.~G. Ramaswamy, C.~Scott, and A.~Tewari.
\newblock Mixture proportion estimation via kernel embedding of distributions.
\newblock In {\em 33$^{~th}$~ICML}, 2016.

\bibitem{rlazerTD}
S.~Reed, H.~Lee, D.~Anguelov, C.~Szegedy, D.~Erhan, and A.~Rabinovich.
\newblock Training deep neural networks on noisy labels with bootstrapping.
\newblock {\em arXiv preprint arXiv:1412.6596}, 2014.

\bibitem{rwCB}
M.~D. Reid and R.~C. Williamson.
\newblock Composite binary losses.
\newblock {\em JMLR}, 11:2387--2422, 2010.

\bibitem{Sanderson:2014}
T.~Sanderson and C.~C.~Scott.
\newblock Class proportion estimation with application to multiclass anomaly
  rejection.
\newblock In {\em AISTATS}, 2014.

\bibitem{sczHI}
F.~Schroff, A.~Criminisi, and A.~Zisserman.
\newblock Harvesting image databases from the web.
\newblock {\em IEEE Transactions on PAMI}, 33(4):754--766, 2011.

\bibitem{Scott:2013}
C.~Scott, G.~Blanchard, and G.~Handy.
\newblock Classification with asymmetric label noise : Consistency and maximal
  denoising.
\newblock In {\em 26$^{~rd}$ COLT}, 2013.

\bibitem{shkssDA}
N.~Srivastava, G.~E. Hinton, A.~Krizhevsky, I.~Sutskever, and R.~Salakhutdinov.
\newblock Dropout: a simple way to prevent neural networks from overfitting.
\newblock {\em JMLR}, 15(1):1929--1958, 2014.

\bibitem{srLS}
G.~Stempfel and L.~Ralaivola.
\newblock Learning {SVM}s from sloppily labeled data.
\newblock In {\em Artificial Neural Networks (ICANN)}, pages 884--893.
  Springer, 2009.

\bibitem{sbpbfTC}
S.~Sukhbaatar, J.~Bruna, M.~Paluri, L.~Bourdev, and R.~Fergus.
\newblock Training convolutional networks with noisy labels.
\newblock In {\em ICLR Workshops}, 2015.

\bibitem{vML}
B.~{van Rooyen}.
\newblock {\em Machine Learning via Transitions}.
\newblock PhD thesis, The Australian National University, 2015.

\bibitem{rmwLW}
B.~{van Rooyen}, A.~K. Menon, and R.~C. Williamson.
\newblock Learning with symmetric label noise: The importance of being
  unhinged.
\newblock In {\em NIPS*29}, 2015.

\bibitem{xxtwLF}
T.~Xiao, T.~Xia, T.~Yang, C.~Huang, and X.~Wang.
\newblock Learning from massive noisy labeled data for image classification.
\newblock In {\em 28$^{th}$ IEEE CVPR}, 2015.

\end{thebibliography}
}

\end{document}